\pdfoutput=1

\documentclass[11pt]{article}

\usepackage[preprint]{acl}
\usepackage{tcolorbox}
\usepackage{enumitem}
\usepackage{algorithm}
\usepackage{algorithmic}

\usepackage{listings}
\usepackage{comment}
\usepackage{booktabs}
\usepackage{makecell}
\usepackage{listings}
\usepackage{xcolor}
\usepackage{listings}
\usepackage{xcolor}
\usepackage{longtable}
\usepackage{multirow}
\usepackage{multirow} 
\definecolor{lightgray}{gray}{0.95}
\usepackage{subcaption}
\usepackage{adjustbox}
\usepackage{amsmath}
\usepackage{siunitx}
\usepackage{amsfonts}

\usepackage{array}
\usepackage[table]{xcolor}
\usepackage{times}
\usepackage{latexsym}

\usepackage{booktabs} 
\usepackage{makecell}

\usepackage{comment}

\usepackage{siunitx} 
\usepackage[colorinlistoftodos]{todonotes}

\usepackage[T1]{fontenc}

\usepackage[utf8]{inputenc}

\usepackage{microtype}

\usepackage{inconsolata}

\usepackage{graphicx}
\tcbuselibrary{listingsutf8}

\lstdefinestyle{jsonstyle}{
  basicstyle=\ttfamily\small,
  breaklines=true,
  breakatwhitespace=false,
  showstringspaces=false,
  columns=flexible,
  keepspaces=true,
  frame=single,
  backgroundcolor=\color{lightgray}
}
\title{When in Doubt, Consult: Expert Debate for Sexism Detection \\ via Confidence-Based Routing}

\author{
  Anwar Alajmi\textsuperscript{1,2}, Gabriele Pergola\textsuperscript{1} \\
  \textsuperscript{1}Department of Computer Science, University of Warwick, Coventry CV4 7AL, UK \\
  \textsuperscript{2}College of Business Studies, Public Authority of Applied Education and Training, Kuwait \\
  \texttt{\{anwar.alajmi, gabriele.pergola.1\}@warwick.ac.uk}
}

\begin{document}
\maketitle
\begin{abstract}
Online sexism increasingly appears in subtle, context-dependent forms that evade traditional detection methods. Its interpretation often depends on overlapping linguistic, psychological, legal, and cultural dimensions, which produce mixed and sometimes contradictory signals in annotated datasets. These inconsistencies, combined with label scarcity and class imbalance, result in unstable decision boundaries and cause fine-tuned models to overlook subtler, underrepresented forms of harm.
To address these challenges, we propose a two-stage framework that unifies (i) targeted training procedures to better regularize supervision to scarce and noisy data with (ii) selective, reasoning-based inference to handle ambiguous or borderline cases. First, we stabilize the training combining class-balanced focal loss, class-aware batching, and post-hoc threshold calibration, strategies for the firs time adapted for this domain to mitigate label imbalance and noisy supervision. Second, we bridge the gap between efficiency and reasoning with a a dynamic routing mechanism that distinguishes between unambiguous instances and complex cases requiring a deliberative process. This reasoning process results in the novel \textit{Collaborative Expert Judgment} (CEJ) module
which prompts multiple personas and consolidates their reasoning through a judge model. Our approach outperforms existing approaches across several public benchmarks, with F1 gains of +4.48\% and +1.30\% on EDOS Tasks A and B, respectively, and a +2.79\% improvement in ICM on EXIST 2025 Task 1.1.

\end{abstract}

\footnotetext[1]{\textbf{Warning:} This paper includes examples that may contain explicit, offensive, or harmful language.
}
\footnotetext[2]{Code will be released upon acceptance.} 

\section{Introduction}

\begin{figure}[h]
  \centering
  \includegraphics[width=\linewidth]{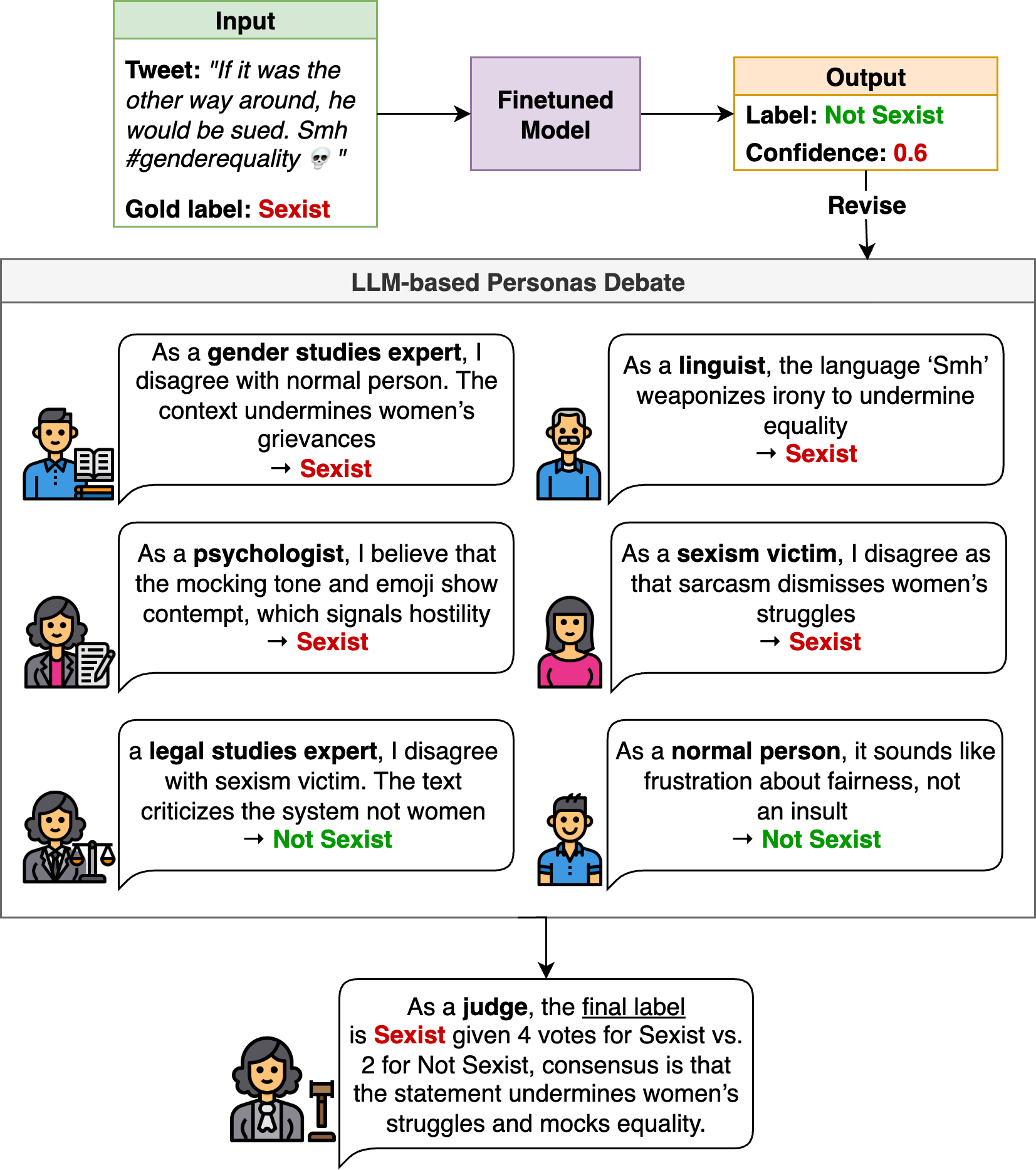}
  \caption{Overview of the proposed framework. \vspace{-12pt}}
  \label{fig:intro}
\end{figure}

Sexist content on social media is increasingly pervasive, often appearing in subtle, sarcastic, or context-dependent forms that evade traditional classification methods. From seemingly innocuous stereotypes to overt hate speech, such content contributes to the emotional and psychological harm that disproportionally affects women \cite{abercrombie-etal-2023-resources}. 

These diverse forms of sexism are shaped by distinct yet overlapping factors: linguistic, psychological, legal, and cultural, among others. As a result, both human annotators and automated systems must contend with mixed and sometimes contradictory signals: a message may appear harmless to a general audience, yet be perceived as harmful by a legal expert, a psychologist, or someone with lived experience of gender-based discrimination \cite{davani2022disagreement}. 

This multidimensionality creates several challenges for annotation and automatic detection. Even the largest and most reliable datasets \cite{edos23, exist25} are typically small, highly imbalanced, and marked by substantial annotator disagreement not only due to ambiguity in individual examples, but also because different annotators may implicitly prioritize different criteria (e.g., linguistic form, psychological impact, cultural normativity, or legal definitions) \cite{khan2025explaining}. These mixed signals can overshadow the representation of less explicit but equally damaging forms of sexism, introducing noise and instability into both the data and the models. At the same time, large language models (LLMs) are pretrained on vast, heterogeneous data sources, absorbing social norms, stereotypes, which are inconsistently represented, and at times in contradiction with human feedback. As a result, fine-tuning these large models on small, noisy sexism datasets is often insufficient to resolve these conflicts, and may inadvertently reinforce dominant or explicit patterns while neglecting subtler cases \cite{khan2025explaining}.
These limitations underscore the need for models that explicitly account for the combined effect of underrepresentation, noise, and conceptual ambiguity in both data and predictions.

To address these challenges, we introduce a two-stage framework that combines (i) targeted training procedures to adapt supervision to scarce and noisy data, with (ii) selective, reasoning-based inference to handle ambiguous cases. This joint design enables the specialized model to make efficient decisions when possible, while engaging with finer contextual reasoning when necessary.
During (i) training, we mitigate long-tailed label distributions and inconsistent annotation through three well-known training strategies. These have never been adapted into an integrated framework for such tasks: the \textit{class-balanced focal loss} (CB-Focal) \citep{ce}, which amplifies the contribution of minority and difficult examples, and the \textit{class-aware batching}, which ensures uniform exposure to underrepresented classes during optimization. We further employ \textit{post-hoc threshold calibration} to stabilize the decision boundaries of the specialist classifier and improve robustness under noisy supervision.
At (ii) inference time, we deploy confidence-aware dynamic routing, which filters the model predictions by difficulty: high-confidence samples are directly classified, while uncertain cases are escalated to a \textit{Collaborative Expert Judgment} (CEJ) module. The CEJ module implements a structured reasoning pipeline composed of multiple \textit{personas}, such as a linguist, a gender expert, a psychologist, and an everyday speaker, each prompted to analyze the same input from distinct perspectives. Their deliberations are subsequently synthesized by a judge model, which integrates these perspectives into a final, interpretable decision (Figure \ref{fig:intro}). The aim is not to faithfully replicate human interaction, but rather to approximate its structure to activate the reasoning and knowledge that instruction-tuned LLMs already encode, yet rarely employ in single-step classification, as evidenced by the limited predictive power of zero-shot baselines \cite{edos23, exist25, khan2025explaining}.

Our experimental evaluation on both the EDOS and EXIST 2025 datasets demonstrates that our methods outperform existing state-of-the-art approaches across almost all tasks. On EXIST 2025 (Task 1.1), the combination of targeted training procedures and confidence-aware routing yields an increase of over 5 F1 points, while also setting new benchmarks on ICM and ICM-Norm metrics. For EDOS, we observe significant improvements in both binary (Task A) and multi-class (Task B) classification: selective escalation with the CEJ module boosts macro F1 by 3.9 points on Task A and ~1 point on Task B. Notably, our approach surpasses previous baselines without employing any ensemble or data augmentation methods.

\noindent Our contributions can be summarized as follows:
\begin{itemize}
\item \textit{A unified framework with selective reasoning capability:} We introduce a unified framework for sexism detection that combines the unmatched efficiency of neural classifiers and targeted training procedures, such as class-balanced focal loss, class-aware batching, and post-hoc threshold calibration, with a reasoning-based inference mechanism that is selectively applied to challenging cases.
\item  \textit{Collaborative Expert Judgment:} We introduce a novel module, \textit{Collaborative Expert Judgment} (CEJ), that leverages multi-persona prompting to resolve ambiguous or borderline cases, enabling the activation of reasoning encoded in instruction-tuned LLMs.
\item \textit{Comprehensive experimental assessment:} We provide a comprehensive analysis of where multi-persona reasoning delivers the greatest benefit, particularly in handling ambiguous and minority cases, and discuss the limitations under severe class imbalance. We demonstrate that integrating these components achieves state-of-the-art results on almost all tasks of both the EXIST 2025 and EDOS datasets.
\end{itemize}

\section{Related Work}\label{sec2}

Our work is aligned with at least two research line on sexism detection and persona-based analyses.

\paragraph{Sexism Classification.} Annotator disagreement, often shaped by cultural perspectives, complicates sexism detection \cite{davani2022disagreement}. Initiatives like the EXIST 2024 shared task encouraged methods robust to inconsistent annotations. Successful approaches leveraged transformer-based models such as DistilBERT \cite{sanh2019distilbert}, DeBERTa \cite{he2020deberta}, RoBERTa \cite{liu2019roberta}, and hybrid ensembles that utilized data augmentation and multi-task learning.
When it comes to LLMs, previous studies have employed adversarial evaluations and data augmentation to enhance detection accuracy. For example, \cite{samory2021call} highlighted standard model vulnerabilities through psychologically grounded adversarial datasets, emphasizing the need for psychologically informed robustness. Similarly, \cite{khan2025explaining} addressed annotator disagreements through definition-driven data augmentation and ensemble methods, improving data-level robustness. 

Beyond explicit abuse, a growing challenge lies in detecting nuanced or implicit forms of hate. The work by \cite{zeng2025sheeps} shows that LLMs often fail to recognize metaphorical expressions of hate speech, highlighting substantial gaps in current moderation capabilities. Their findings emphasize that even state-of-the-art models struggle when hateful intent is obscured through indirect language, a phenomenon that is particularly relevant for sexist discourse. Building on this, the authors in \cite{fasching2025model} demonstrate that hate speech detection outcomes vary significantly across LLM-based systems, underscoring how moderation decisions remain model-dependent and inconsistent. Our work directly responds to these challenges by integrating confidence-aware routing and multi-expert arbitration, ensuring that inconsistent or ambiguous cases are addressed through structured reasoning and synthesis.

\paragraph{Persona-based Approaches.} More recent work integrates multi-agent architectures, expert simulations, and human-in-the-loop reasoning to improve transparency and social norm alignment. While expert persona-driven prompting strategies, such as those in \cite{xu2023expertprompting,long2024multi}, have advanced interpretability, our method surpasses static personas by introducing dynamic, structured, and confidence-aware interactions among multiple expert roles, capturing richer contextual nuances. Moreover, our framework integrates a dedicated judge model to arbitrate conflicts, providing nuanced decisions beyond simple voting mechanisms.

Emerging research demonstrated the effectiveness of multi-expert deliberation in LLM-based decision-making. \cite{lu-etal-2024-triageagent} introduce a multi-agent framework for clinical triage where LLM agents assume distinct roles, engage in multi-round discussions with self-confidence scoring, and iteratively revise their assessments until reaching consensus. Similarly, \cite{long2024multi} simulate multiple domain experts within a single LLM, aggregate their responses, and select the best one. Both approaches show that structured discussion among diverse perspectives yields more reliable outputs than single-viewpoint reasoning, as cross-perspective validation mitigates individual biases and enables principled conflict resolution before final judgment. Our framework extends this paradigm to sexism detection, where the inherently subjective nature of the task makes perspectival diversity essential.

Another notable study that effectively leverages expert agents is Expert-Token-Routing \cite{chai2024expert}, which adopts a router-based approach similar to that of \cite{jiang2023llm}. In this method, a router model is trained to direct each question or instruction to a specialized expert, where the expert LLM is responsible for generating the next token. Their approach encodes expert LLMs as special tokens within the vocabulary of the meta-LLM, drawing inspiration from ToolkenGPT \cite{hao2023toolkengpt}.

Our progressive prompt refinement aligns with iterative human-in-the-loop methods discussed in \cite{shah2025prompt}, systematically enhancing logical coherence and prompting effectiveness. In addition, we preserve the logical structure and enhance the prompt formatting for effective model reasoning as it is shown by \cite{li2025llms}'s experiments.

\begin{figure*}[ht]
  \centering
  \includegraphics[width=.95\linewidth]{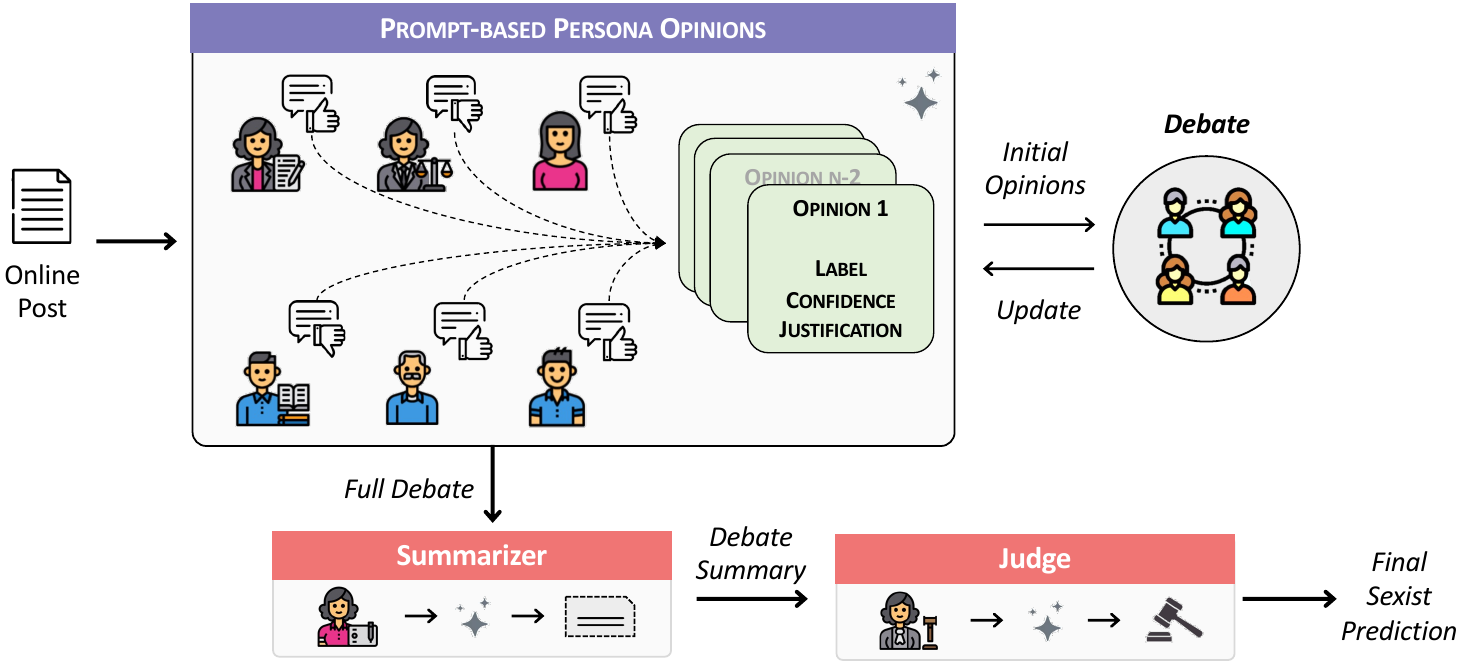}
  \caption{CEJ framework, where a collaborative reasoning discussion among personas with refinement is generated before the judge model makes a definitive decision.}
  \label{fig:expert2}
\end{figure*}

\section{Methodology}\label{sec3}
We introduce a two-stage framework for sexism detection. The two stages are based on the core premise that not all instances require complex reasoning. And while many can be better resolved by the specialist model, forcing indiscriminately an immediate decision on ambiguous cases leads to errors. We therefore implement a selective classification mechanism  that acts as a gatekeeper between the (i) specialised classifier and the (ii) reasoning module. The specialized model is fine-tuned moving beyond standard cross-entropy minimization by integrating three regularization strategies designed to stabilize learning on the unstable decision boundaries caused by the severe class imbalance and annotator disagreement. Then, we couple this mechanism with a confidence-aware routing mechanism which selects low-confidence instances the needs to be escalated to the \textit{Collaborative Expert Judgment} (CEJ) module, where multiple personas are designed to reason, deliberate, and produce the final decision (Figure \ref{fig:expert2}). 


\subsection{Domain-Tuned Model}
We first fine-tune a small instruction-tuned LLM as a domain-specialized classifier. To handle severe class imbalance, we integrate three complementary strategies: the \textit{Class-Balanced Cross-Entropy} (CB-CE) \citep{ce}, the \textit{Focal Loss} \citep{ce}, and the \textit{Class-Aware Batch Sampling} (CAB) \citep{henning23}. 
With the CB-CE, rather than weighing samples by raw frequency, which can be misleading in long-tailed distributions, we calculate the effective number of samples to capture the diminishing marginal benefit of additional samples. This is combined with the \textit{Focal Loss} objective, which dynamically down-weights easy negatives and focuses the gradient on hard examples, i.e., those that are misclassified or ambiguous. 
We further ensure consistent exposure to underrepresented categories through CAB, decoupling batch composition from the dataset's natural distribution, so that gradient updates are not dominated by majority class signals.
Finally, we apply post-hoc threshold calibration via temperature scaling \citep{guo2017calibration}, so that the model’s output probabilities accurately align with its epistemic uncertainty, preventing it from assigning high confidence to ambiguous cases where it lacks evidence.
For binary classification, we apply temperature scaling and tune a decision threshold on the development set; for multi-class classification, we use CB-Focal loss given the more severe imbalance.


\subsection{Confidence-Aware Routing}
Following the initial classification, we employ a routing mechanism to handle instances where the specialised model exhibits high epistemic uncertainty, i.e., ambiguous examples. To systematically identify these uncertain predictions, we utilize a selective classification rule \citep{elyaniv2010foundations}. For an input $x$, the model produces a predicted label $\hat{y}_s$ with the associated confidence score $c_s(x)$:
\begin{align}
\hat{y}_s &= \arg\max_k p_\theta(y{=}k \mid x)\\
c_s(x) &= \max_k p_\theta(y{=}k \mid x)
\end{align}
If $c_s(x) \ge \tau_{\text{conf}}$, we accept $\hat{y}_s$; otherwise, the input is routed to the CEJ module. 

In multi-class settings, reliance on raw confidence scores alone is insufficient because a model may often assign a relatively high probability to the top class while remaining deeply conflicted between two competing categories (e.g., \textit{Derogation} vs. \textit{Animosity}). We thus define the margin between the two most probable labels as
$m(x) = p_{(1)}(x) - p_{(2)}(x)$
where $p_{(1)}(x)$ and $p_{(2)}(x)$ denote the largest and second-largest posterior probabilities. An instance is then routed to CEJ if:
\begin{equation}
\big(c_s(x) < \tau_{\text{conf}}\big) \;\land\; \big(m(x) < \tau_{\text{margin}}\big) \footnote{Both thresholds are tuned jointly on development data to maximize macro-F1 while minimizing unnecessary escalation.}
\end{equation}

\subsection{Collaborative Expert Judgment (CEJ)}

By acting as a gatekeeper, the aforementioned routing mechanism identify the instances where statistical pattern matching fails, and ensures that only the subset of difficult cases triggers the accurate but computationally expensive Collaborative Expert Judgment (CEJ) module. 
This reasoning module is based on prior research on abusive and sexist language which has widely highlighted the importance of combining expert knowledge with lived experience \cite{sap2020socialbiasframes, davani2022disagreement}. We therefore structure CEJ as a multi-perspective reasoning module based on a set of \textit{personas} that act as distinct interpretive lenses. The `expert' personas, such as \textit{linguists}, \textit{psychologists}, \textit{gender studies scholars}, and \textit{legal professionals} are chosen based on the literature showing that they are the main critical perspectives on language use, structural bias, emotional impact, and legal norms \cite{lazar2005feminist, vidgen2019challenges}. Yet, to still capture non-expert and experiential perspectives, we incorporate a \textit{sexism victim} persona, reflecting direct experience of harm, and a \textit{layperson} persona, capturing everyday interpretations often responsible for annotator disagreement (Table~\ref{tab:unified_personas}).

It is worth noting that the aim of this multi-persona prompting is not to simulate faithful and exhaustive expert panels, but rather to elicit and organize the knowledge already encoded within LLMs. This combined approaches induces the CEJ module to trigger and controll the LLMs' domain understanding that could otherwise remain latent in single-step classification.


\begin{table*}[t]
\centering
\begin{adjustbox}{width=\textwidth}
\setlength{\tabcolsep}{4pt}
\begin{tabular}{@{}l@{} c@{} l@{} l c c c c@{}}
\toprule
\textbf{\#} & \textbf{Variant \;} & \textbf{Model} & \textbf{ICM} & \textbf{ICM-Norm} & \textbf{F1\_YES} & \textbf{F1} \\
\midrule
&&\multicolumn{5}{@{}l}{\textbf{Zero-shot Baselines}} \\
1 &  & \textsc{GPT-5.2}\textsuperscript{$\dagger$}         & 0.3408 & 0.6705 & 0.7043 & 0.7349 \\
2 &  & \textsc{Qwen2.5-72B-Instruct}          & 0.3765 & 0.6892 & 0.7112 & 0.7384 \\
3 &  & \textsc{LLaMA-3.3-70B-Instruct}        & 0.3002 & 0.6509 & 0.6868 & 0.7145 \\
4 &  & \textsc{Cogito-70B}                    & 0.3656 & 0.6837 & 0.7109 & 0.7348 \\
5 &  & \textsc{Cogito-70B} (reasoning)        & 0.3772 & 0.6896 & 0.7143 & 0.7384 \\

\midrule
&&\multicolumn{5}{@{}l}{\textbf{EXIST 2025 Task 1.1 Leaderboard}} \\
6  &  & BERT \cite{nowakowski2025exist}                             & 0.5727 & 0.7878 & 0.7802 & -- \\
7  &  & XLM-RoBERTa \cite{pan2025exist}                              & 0.5799 & 0.7915 & 0.7824 & -- \\
8  &  & Dual-Transformer Fusion Network \cite{khan2025exist}         & 0.5806 & 0.7918 & 0.7837 & -- \\
9  &  & DeepSeek-R1-Distill-Llama-8B \cite{villarreal2025exist} \;     & 0.6127 & 0.8079 & 0.7945 & -- \\
10  &  & Ensemble Approach \cite{c:22}                                & 0.6249 & 0.8141 & 0.7991 & -- \\
11 &  & XLM-RoBERTa \cite{c:21}                                      & 0.6297 & 0.8165 & 0.7996 & -- \\
12 &  & LLaMA-3.1-8B-Instruct \cite{tian2025marioexist2025simple}    & 0.6774 & 0.8405 & 0.8167 & -- \\
\midrule
&&\multicolumn{5}{@{}l}{\textbf{Ours}} \\
13 & $\mathcal{C}_{1}$ & \textsc{LLaMA-3.2-3B}                                         & 0.5709 & 0.7913 & 0.7596 & 0.7934 \\
14 & $\mathcal{C}_{2}$ & $\mathcal{C}_{1}$ + Task-specific optimizations               & 0.6595 & 0.8315 & 0.8089 & 0.8257 \\
15 & $\mathcal{C}_{3}$ & $\mathcal{C}_{2}$ + Routing to CEJ$_{\mathrm{Q}\to\mathrm{C}}$ & \underline{0.6831} & \underline{0.8433} & \underline{0.8191} & \underline{0.8348} \\
16 & $\mathcal{C}_{4}$ & $\mathcal{C}_{2}$ + Routing to CEJ$_{\mathrm{L}\to\mathrm{C}}$ & \textbf{0.6963} & \textbf{0.8500} & \textbf{0.8233} & \textbf{0.8389} \\
\bottomrule
\end{tabular}
\end{adjustbox}
\caption{Results on the EXIST 2025 Task~1.1 test set with 
zero-shot baselines, leaderboard submissions, and our approach. 
"Q" and "L" denote the models used for personas debate (\textsc{Qwen2.5-72B-Instruct} and \textsc{LLaMA-3.3-70B}, respectively), and "C" refers to the judge model \textsc{Cogito-70B} (reasoning mode). 
\textbf{Bold} = best score, \underline{underline} = runner-up. \textsuperscript{$\dagger$}Evaluated on the development set; the official evaluation platform (\url{https://evall.uned.es/}) was unavailable at time of submission.}

\label{tab:ensemble-performance}
\end{table*}

\subsubsection{Structured Debate and Judge} 
The CEJ process comprises four sequential prompt-based stages, designed to elicit, interrogate, and synthesize domain-relevant knowledge:\\
\noindent \textbf{Initial Opinions.} Each persona independently analyzes the input, providing an initial classification, justification, with a confidence score. This initial stage establishes diverse baseline perspectives. \\    
\noindent \textbf{Structured Debate.}
Personas are subsequently exposed to all initial opinions and begging a critical evaluation of their peers’ reasoning. During this deliberation, each persona must: (i) engage with at least one other perspective, agreeing or disagreeing while offering supporting rationale; (ii) revise their own stance if confronted with compelling counter-arguments; and (iii) reassess the input through the interpretive lens of other personas, re-evaluating both intent and target. This process results in a revised classification, an updated justification, and an adjusted confidence score for each persona. \\
%
\noindent \textbf{Summarization.} Then, a dedicated summarization agent condenses the full debate into a concise synthesis highlighting the main arguments, points of consensus, and unresolved disagreements.\\
\noindent \textbf{Final Judgment.} Finally, the \textit{judge} model synthesizes all available evidence, i.e., the original input, both initial and revised persona opinions, and the debate summary, to produce a final adjudication. Prompted to act impartially, the judge delivers a final classification, comprehensive justification, and a confidence score.


\begin{table*}[t]
\small
\centering
\begin{adjustbox}{width=0.97\textwidth}
\setlength{\tabcolsep}{17pt}
\begin{tabular}{@{}l@{} c@{} l@{} l c c c@{}}
\toprule
\textbf{\#} & \textbf{Variant \;} &  & \textbf{Model} & \textbf{Task A} & \textbf{Task B} & \textbf{Task C} \\
\midrule
&&&\multicolumn{4}{@{}l}{\textbf{Zero-shot Baselines}} \\
1  &  &  & \textsc{GPT-5.2}  & 0.7306 & 0.4543 & 0.3381 \\
2  &  &  & \textsc{Qwen-2.5-72B-Instruct}           & 0.6190 & 0.4246 & 0.2576 \\
3  &  &  & \textsc{LLaMA-3.3-70B-Instruct} & 0.6421 & 0.4297 & 0.2775 \\
4  &  &  & \textsc{Cogito-70B}             & 0.6106 & 0.4214 & 0.2683 \\
5  &  &  & \textsc{Cogito-70B} (reasoning) & 0.6332 & 0.4325 & 0.2826 \\
\midrule
&&&\multicolumn{4}{@{}l}{\textbf{SemEval 2023 Task 10.A Leaderboard}} \\
6  &  &  & DeBERTa-v3-large + twHIN-BERT-large \cite{zhou-2023-pinganlifeinsurance} & 0.8746 & -- & -- \\
7  &  &  & RoBERTa-Large + ELECTRA                                                    & 0.8740 & 0.7203 & 0.5487 \\
8  &  &  & DeBERTa Ensemble                                                           & 0.8740 & -- & -- \\
\midrule
&&&\multicolumn{4}{@{}l}{\textbf{Data Augmentation and Ensemble Methods}} \\
9  &  &  & SEFM \cite{zhong-etal-2023-uirisc}            & 0.8538 & 0.6619 & 0.4641 \\
10  &  &  & QCon \cite{feely-etal-2023-qcon}              & 0.8400 & 0.6400 & 0.4700 \\
11 &  &  & HULAT \cite{segura-bedmar-2023-hulat-semeval} & 0.8298 & 0.5877 & 0.4458 \\
12 &  &  & CSE \cite{khan2025explaining}                 & 0.8819 & 0.7243 & 0.5639 \\
13 &  &  & DDA \cite{khan2025explaining}                 & 0.8769 & 0.7277 & \textbf{0.6018} \\
14 &  &  & PaLM Ensemble                                 & --     & \underline{0.7326} & -- \\
\midrule
&&&\multicolumn{4}{@{}l}{\textbf{Ours}} \\
15 & $\mathcal{C}_{1}$ &  & \textsc{LLaMA-3.2-3B}                           & 0.7824 & 0.5941 & 0.3717 \\
16 & $\mathcal{C}_{2}$ &  & $\mathcal{C}_{1}$ + Task-specific regulations & 0.8986 & 0.6279 & 0.4189 \\
17 & $\mathcal{C}_{3}$ &  & $\mathcal{C}_{2}$ + Routing to CEJ$_{\mathrm{Q}\to\mathrm{C}}$ & \underline{0.9195} & 0.7324 & 0.5842 \\
18 & $\mathcal{C}_{4}$ &  & $\mathcal{C}_{2}$ + Routing to CEJ$_{\mathrm{L}\to\mathrm{C}}$ & \textbf{0.9214} & \textbf{0.7421} & \underline{0.5904} \\
\bottomrule
\end{tabular}
\end{adjustbox}
\caption{Macro-F1 scores on the EDOS datasets with zero-shot baselines, leaderboard references, augmentation/ensemble methods, and our approach. \textbf{Bold} = best score, \underline{underline} = runner-up.}
\label{tab:edos}
\end{table*}

\section{Experiments} \label{sec4}
We proceed describing the datasets, evaluation metrics, and results used to assess the proposed framework.

\subsection{Experimental Setup}
\noindent \textbf{Datasets.} We evaluate our methods on two widely used benchmarks for sexism detection: the EXIST 2025 Tweets dataset \cite{exist25} and the EDOS dataset \cite{edos23}. The EXIST dataset incorporates annually updated social media posts, capturing emerging sexist language, evolving slang, and annotator disagreements that mirror ambiguities in human judgment. The EDOS dataset offers a complementary perspective with high quality fine-grained annotations at multiple levels of granularity, organized across three hierarchical classification tasks (see Appendix~\ref{dsandtasks}).

\noindent \textbf{Evaluation Metrics.} Aligned with the official setup, we use the \textit{Evaluate ALL 2.0} tool \cite{evall2025} for scoring under the EXIST 2025 Task 1.1 (Hard-Hard) configuration. The evaluation metrics include: (i) Inter-Consistency Measure (ICM) \cite{amigo2022evaluating}, which quantifies agreement between predicted and reference annotations by combining intra-system consistency and penalizing overlap; (ii) ICM-Norm, a normalized version of ICM scaled to $[0,1]$; and the (iii) F1 score.
As for the EDOS dataset, we follow the public evaluation protocol, reporting results for the Macro-F1 score, the unweighted mean of the F1-scores computed independently for each class across task A, B, and C (Appendix \ref{dsandtasks}).

\noindent \textbf{Baselines.}
We evaluate our proposed framework against both zero-shot and fine-tuned LLM baselines. It worth noting that we focus on state-of-the-art approaches for sexism detection rather than aiming at an exhaustive exploration of all available LLMs, which would be beyond the scope of this work. 
As specialized models, we employ \textsc{LLaMA-3.2-3B}, adapted via LoRA fine-tuning \citep{hu2021lora} on both the EDOS and EXIST training datasets. Full task-specific training information is available in Appendix \ref{training}. 
To evaluate instruction-following and reasoning abilities in a zero-shot setting, we test \textsc{Qwen2.5-72B-Instruct} \cite{qwen2025qwen25technicalreport}, \textsc{LLaMA-3.3-70B-Instruct} \cite{grattafiori2024llama}, \textsc{GPT-5.2} \cite{openai2025gpt5}, and \textsc{Cogito-70B} \cite{deepcogito_cogito_v2} using consistent classification prompts (Appendix~\ref{zeroshot}). Pre-trained checkpoints were obtained from HuggingFace \cite{huggingface}. All experiments were conducted locally through LangChain \cite{langchain} and Ollama \cite{ollama}, on a system equipped with 3 Nvidia A100 GPUs.

\subsection{Results}


\noindent \textbf{EXIST Dataset.} Results for the binary classification task 1.1 of EXIST 2025 are reported in Table~\ref{tab:ensemble-performance}. The first group of models comprises LLMs evaluated in a zero-shot setting. Among these, \textsc{Cogito-70B} (reasoning mode) achieves the highest F1\_YES (0.7143), outperforming by little \textsc{Qwen2.5-72B-Instruct} (0.7112), standard \textsc{Cogito-70B} (0.7109), and \textsc{GPT-5.2} (0.7043). Overall, all zero-shot models with simple prompts still underperform smaller yet specialised approaches, reflecting still the limitations of LLMs prompted with simple requests.
The second group reports public leaderboard systems submitted to EXIST 2025 Task 1.1. Here, LLaMA-3.1-8B-Instruct \cite{tian2025marioexist2025simple} leads in both ICM and F1\_YES. Systems based BERT-variants still shows competitive results compare to larger architectures with simple prompts.

The bottom of the Table~\ref{tab:ensemble-performance} details the results of our routing framework. The initial specialised model, \textsc{LLaMA-3.2-3B}, achieves performance on par with lower-ranking leaderboard models (F1\_YES: 0.7596). Applying simple but targeted regularization, indicated by $\mathcal{C}_2$ (multi-dataset training, class-balanced loss, threshold calibration), yields already substantial improvements: 0.8089 F1\_YES and 0.8257 macro-F1, with outperforming several leaderboard submissions.
Further gains are realized through the dynamic routing based on confidence calibrations. Configuration $\mathcal{C}_3$ integrates \textsc{Qwen2.5-72B-Instruct} for persona-based debate, reaching 0.8191 F1\_YES, while $\mathcal{C}_4$, leveraging \textsc{LLaMA-3.3-70B} for personas and \textsc{Cogito-70B} as judge, achieves the highest scores across all reported metrics. This configuration, achievement state-of-the-art peformance on the task, demonstrates the importance of controlling different LLMs to leverage wider capabilities for the task.  

\noindent \textbf{EDOS Dataset.} For the EDOS dataset, we report in Table~\ref{tab:edos} the Macro-F1 across the three classification tasks of increasing granularity. 

The first group reports the results of LLMs evaluated in a zero-shot setting. Performance remains limited across all tasks when using simple prompts, with \textsc{GPT-5.2} achiving the highest performance across all tasks, followed by the open-weight \textsc{LLaMA-3.3-70B-Instruct}. \textsc{Cogito-70B} in reasoning mode consistently outperforms its standard variant. However, all zero-shot models particularly struggle with fine-grained classification (Task C).
Among official leaderboard systems from SemEval 2023 Task 10.A, ensemble methods based on DeBERTa and RoBERTa architectures are also in this case strong baselines.
The next group reports systems employing data augmentation, model ensembling, or both. CSE \cite{khan2025explaining} and DDA \cite{khan2025explaining} achieve notable gains: CSE reaches 0.8819 on Task A and 0.5639 on Task C, motivating the data expansion and leveraging perspectives from multiple models.

Our approach in the base configuration, $\mathcal{C}_1$, i.e., fine-tuned \textsc{LLaMA-3.2-3B}, achieves competitive results for binary and multi-class tasks despite its simplicity, but falls short on fine-grained tasks. Introducing targeted training regularization also here, $\mathcal{C}_2$, yields again marked improvements, outperforming the zero-shot baselines and most of the specilised systems.
Finally, confidence-aware escalation to the CEJ module provides substantial gains. Configuration $\mathcal{C}_3$, which employs \textsc{Qwen2.5-72B-Instruct} for persona-based debate, achieves 0.9195 on Task A and 0.7324 on Task B, with the strongest results are obtained by $\mathcal{C}_4$, where \textsc{LLaMA-3.3-70B} is used for the persona debate and \textsc{Cogito-70B} as the judge. Both configurations outperform all previous systems on binary and multi-class classification on task A and B. On Task C, both routing variants remain competitive, 
though they struggle compare to the DDA data augmentation method. This gap can be attributed to the cascading effect of baseline limitations: despite employing task-specific optimization techniques, ($\mathcal{C}_2$) exhibits progressive macro-F1 score degradation as class granularity increases (0.8986 $\rightarrow$ 0.6279 $\rightarrow$ 0.4189 across Tasks A, B, and C), indicating that these optimization techniques only partially mitigate the challenge of learning 11 fine-grained categories with severe class imbalances. 
In contrast, DDA addresses data scarcity directly by synthesizing training examples that populate underrepresented categories, providing a stronger foundation that inference-level refinement alone cannot replicate.

\begin{table}[t]
\centering
\small
\resizebox{\columnwidth}{!}{%
\begin{tabular}{lccccc}
\toprule
\textbf{Category} & \textbf{n} & $\mathcal{C}_{2}$ & $\mathcal{C}_{3}$ & $\mathcal{C}_{4}$ & \textbf{Gain} \\
\midrule
\multicolumn{6}{@{}l}{\textit{Task B}} \\
1. Threats/Harm     & 89  & .681 & .848 & .857 & +17.6 \\
2. Derogation       & 454 & .727 & .608 & .745 & +1.9 \\
3. Animosity        & 333 & .563 & .663 & .519 & +10.1 \\
4. Prejudiced Disc. & 94  & .541 & .811 & .848 & +30.7 \\
\midrule
\multicolumn{6}{@{}l}{\textit{Task C}} \\
Cat. 1.1 & 16  & .217 & .300 & .250 & +8.3 \\
Cat. 1.2 & 73  & .626 & .771 & .767 & +14.5 \\
Cat. 2.1 & 205 & .568 & .670 & .646 & +10.2 \\
Cat. 2.2 & 192 & .537 & .571 & .561 & +3.4 \\
Cat. 2.3 & 57  & .426 & .514 & .528 & +10.2 \\
Cat. 3.1 & 182 & .700 & .739 & .728 & +3.9 \\
Cat. 3.2 & 119 & .556 & .735 & .724 & +18.0 \\
Cat. 3.3 & 18  & .231 & .973 & .973 & +74.2 \\
Cat. 3.4 & 14  & .100 & .933 & .933 & +83.3 \\
Cat. 4.1 & 21  & .408 & .069 & .188 & $-$22.1 \\
Cat. 4.2 & 73  & .238 & .152 & .198 & $-$4.1 \\
\bottomrule
\end{tabular}%
}
\caption{Class-wise F1 on EDOS Tasks B and C. 
$\mathcal{C}_2$: domain-tuned model; 
$\mathcal{C}_3$: CEJ routing (Qwen); 
$\mathcal{C}_4$: CEJ routing (LLaMA). 
Gain: improvement from $\mathcal{C}_2$ to best CEJ configuration (percentage points).}
\label{tab:classwise_performance}
\end{table}

Table~\ref{tab:classwise_performance} presents class-wise F1 scores for 
Tasks B and C. CEJ shows an inverse relationship between class frequency 
and effectiveness: rarer categories benefit most, with gains of +30.7 
(\textit{Prejudiced Discussions}, $n$=94) and +17.6 (\textit{Threats/Harm}, 
$n$=89) in Task B. This pattern holds in Task C, 
where the smallest classes 3.3 ($n$=18) and 3.4 ($n$=14) achieve gains of 
+74.2 and +83.3 points. This suggests that multi-perspective debate compensates for representation bias in underrepresented categories.

However, CEJ degrades performance for categories 4.1 ($-$22.1) and 4.2 
($-$4.1). These subcategories, 4.1 (\textit{Supporting mistreatment of 
individual women}) and 4.2 (\textit{Supporting systemic discrimination}), 
require subtle distinctions: 4.1 separates victim-blaming from neutral 
risk statements, while 4.2 separates endorsement of discrimination from 
good-faith debate about the progress of women's rights. The domain-tuned model learns these annotation-specific boundaries 
directly from training data. CEJ, however, introduces multiple interpretive 
perspectives that may not align with these boundaries, adding variance where 
the task requires consistency with annotator conventions.

\begin{figure}[tb]
  \centering
  \includegraphics[width=\linewidth]{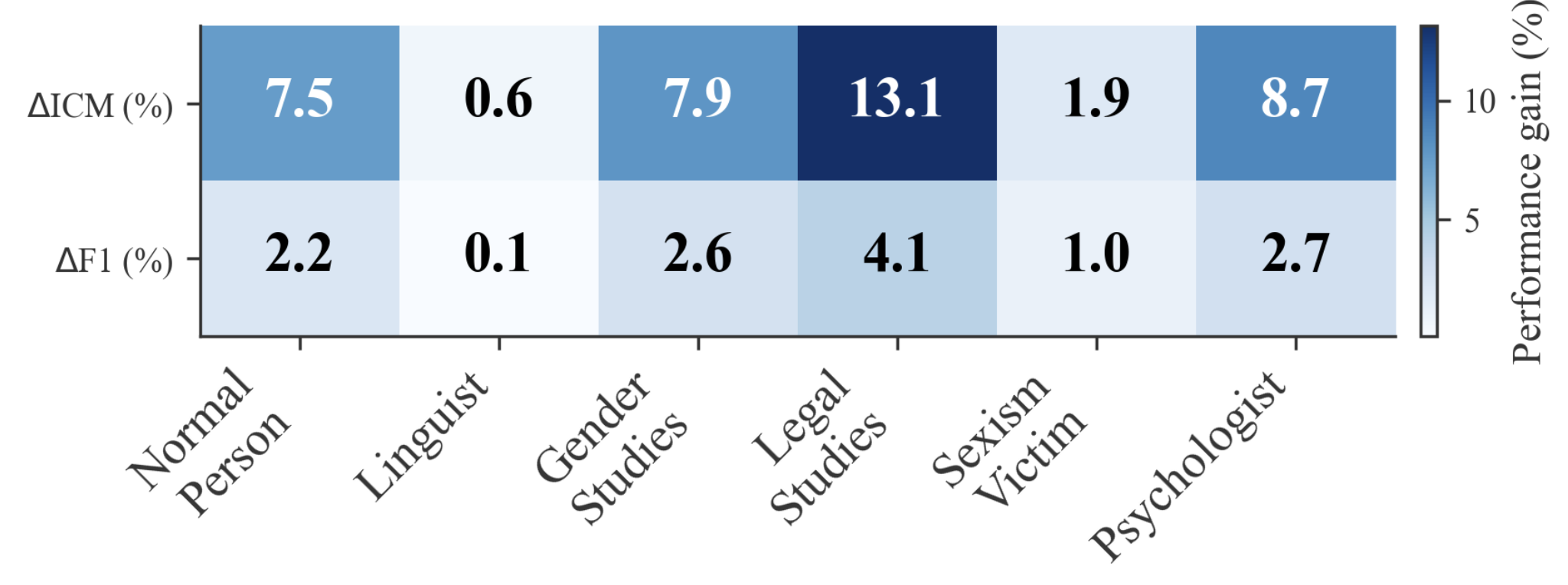}
  \caption{Performance gain after the debate phase.}
  \label{fig:persona_heatmap}
\end{figure}

Figure~\ref{fig:persona_heatmap} presents per-persona gains after the debate 
phase. The legal studies persona shows the largest improvement 
(+13.1\% ICM, +4.1\% F1), as legal assessments inherently require synthesizing 
intent, context, and harm—dimensions enriched through cross-persona 
deliberation. Psychologist (+8.7\%) and gender studies expert 
(+7.9\%) also benefit substantially.

\paragraph{Ablation Study}
Table~\ref{tab:prompt_ablation} presents the prompt ablation results. 
The full prompt configuration (\(\mathcal{P}_5\)) consistently outperforms 
ablated variants across all personas. The largest degradations occur when 
removing sexism definitions (\(\mathcal{P}_2\)) or persona-specific context 
(\(\mathcal{P}_1\)), with the \textit{Psychologist} exhibiting the greatest 
sensitivity (+19.3\% ICM gain from \(\mathcal{P}_1\) to \(\mathcal{P}_5\)). 
Notably, ICM and ICM-Norm improvements exceed F1 gains, indicating enhanced 
alignment with human annotation distributions rather than mere accuracy.
\begin{table}[t]
\centering
\footnotesize
\setlength{\tabcolsep}{2mm}
\begin{tabular}{@{}lccccc@{}}
\toprule
\textbf{Persona} & 
$\mathcal{P}_1$ & 
$\mathcal{P}_2$ & 
$\mathcal{P}_3$ & 
$\mathcal{P}_4$ & 
$\mathcal{P}_5$ \\
\midrule
Normal Person         & .442 & .459 & .561 & .593 & \textbf{.598} \\
Linguist              & .439 & .446 & .584 & .563 & \textbf{.586} \\
Gender Studies Expert & .433 & .443 & .577 & .576 & \textbf{.583} \\
Legal Studies Expert  & .466 & .453 & .558 & .566 & \textbf{.573} \\
Sexism Victim         & .433 & .469 & .568 & .571 & \textbf{.592} \\
Psychologist          & .406 & .446 & .574 & .593 & \textbf{.599} \\
\bottomrule
\end{tabular}
\caption{Prompt ablation (ICM scores, \textsc{LLaMA-3.3-70B}). 
$\mathcal{P}_1$: baseline few-shot with role identity. 
$\mathcal{P}_2$: specialized expert persona. 
$\mathcal{P}_3$: formal sexism definitions. 
$\mathcal{P}_4$: nuanced multilingual examples. 
$\mathcal{P}_5$: refined guidelines from error analysis.
}
\label{tab:prompt_ablation}
\end{table}

\section{Conclusion}\label{sec5}
In this work we addressed some of the structural issues in sexism detection, including data scarcity, class imbalance, and annotation noise; factors that often undermine the reliability of standard fine-tuning. We proposed a unified framework that augments targeted training with confidence-calibrated routing and multi-persona deliberation. Through a combination of robust loss functions, balanced batching, and post-hoc calibration, our system improves specialist reliability, while selectively escalating ambiguous cases to a structured, multi-perspective judgment module, CEJ.
Evaluations on EXIST 2025 and EDOS show that this approach not only advances state-of-the-art performance, but also proves especially effective on cases marked by ambiguity or underrepresentation.


\section*{Limitations}
While our frameworks demonstrate strong performance and interpretability through expert-guided multi-persona reasoning, several limitations remain. Our approaches utilize prompt engineering and LLM-generated content, which can reflect biases present in the models' pretraining data. Additionally, the scalability of multi-agent interactions poses a practical constraint: as the number of simulated personas increases, so does computational cost, which may hinder applicability in real-time or resource-limited settings.

Moreover, our system's performance degrades with highly granular classification 
schemes. The domain-tuned baseline deteriorates with increasing class 
granularity and severe class imbalances, and this limitation propagates 
through the routing pipeline. This suggests that reasoning-based refinement 
is better suited for binary or moderate-granularity tasks, while extreme 
class fragmentation with heavy imbalances may require complementary data-level 
interventions such as the DDA method. Importantly, these approaches are not 
mutually exclusive; CEJ could potentially be combined with data augmentation 
strategies to leverage both enriched training distributions and structured 
inference-time reasoning. We leave exploration of such hybrid architectures 
to future work.


\bibliography{custom}

\newpage

\appendix

\setcounter{table}{0}
\renewcommand{\thetable}{A\arabic{table}}
\setcounter{figure}{0}
\renewcommand{\thefigure}{A\arabic{figure}}

\clearpage
\section{Appendix}\label{Appendix}
\subsection{Datasets Details and Task Structure}\label{dsandtasks}

The EDOS dataset (SemEval-2023 Task 10) consists of 14,000 training, 2,000 development, and 4,000 test examples for Subtask A (binary sexism detection) including 970 sexist examples for Subtasks B and C. It defines a three-level hierarchical classification problem:

\begin{itemize}
    \item \textbf{Task A}: \emph{Binary classification} — distinguishing between \textit{sexist} and \textit{not sexist} texts.
    
    \item \textbf{Task B}: \emph{Four-way categorization of sexist content}, assigning each sexist instance to one of the following categories:
    \begin{enumerate}
        \item Threats, plans to harm and incitement
        \item Derogation
        \item Animosity
        \item Prejudiced discussions
    \end{enumerate}
    
    \item \textbf{Task C}: \emph{Fine-grained subcategorization}, where each sexist instance is further assigned to one of the following subcategories:
    \begin{enumerate}
        \item[1.1] Threats of harm
        \item[1.2] Incitement and encouragement of harm
        \item[2.1] Descriptive attacks
        \item[2.2] Aggressive and emotive attacks
        \item[2.3] Dehumanising attacks \& overt sexual objectification
        \item[3.1] Casual use of gendered slurs, profanities, and insults
        \item[3.2] Immutable gender differences and gender stereotypes
        \item[3.3] Backhanded gendered compliments
        \item[3.4] Condescending explanations or unwelcome advice
        \item[4.1] Supporting mistreatment of individual women
        \item[4.2] Supporting systemic discrimination against women as a group
    \end{enumerate}
\end{itemize}

For Tasks B and C, we train on filtered subsets where \texttt{label\_sexist} = ``Sexist'' from Task A predictions. Each task uses a separate specialist model with task-specific label mappings.

In addition, we evaluate on the EXIST 2025 dataset, which focuses on sexism detection in a cross-lingual, cross-platform setting. It contains 10,034 binary-labeled samples (\texttt{YES} or \texttt{NO}) in English and Spanish. The dataset includes 6,920 training samples (3,260 English, 3,660 Spanish), 1,038 development samples (489 English, 549 Spanish), and 2,076 test samples (978 English, 1,098 Spanish). The task is structured as follows:
\begin{itemize}
    \item \textbf{Task 1.1 (EXIST)}: Binary classification of social media posts as \emph{Sexist} vs. \emph{Not Sexist}. 
    This task is analogous to EDOS Task~A, but differs in scale, language diversity (English and Spanish), and domain coverage (Twitter and Gab), thereby presenting a more challenging generalization setting.
\end{itemize}

\subsection{Training Details}\label{training}
\subsubsection{Shared Training Infrastructure Details (Binary and Multi-Class)}\label{shared}
We employ \textsc{\textsc{LLaMA-3.2-3B}} as our base model with 4-bit quantization to enable efficient fine-tuning on consumer hardware while maintaining model quality. The quantization configuration is as follows:
\begin{itemize}
    \item \textbf{Quantization type:} NF4 (Normal Float 4-bit)
    \item \textbf{Double quantization:} Enabled
    \item \textbf{Compute dtype:} bfloat16 (if supported) or float16
\end{itemize}

LoRA adapters are applied to all attention and feed-forward projection layers (\texttt{q\_proj}, \texttt{k\_proj}, \texttt{v\_proj}, \texttt{o\_proj}, \texttt{gate\_proj}, \texttt{up\_proj}, \texttt{down\_proj}). Configuration varies by task (see Table~\ref{tab:hyperparameters}). This reduces trainable parameters to $<1\%$ of the full model while maintaining competitive performance across all tasks.

\subsubsection{Binary Classification Optimizations}\label{binary}
Class-balanced cross-entropy (CB-CE) is employed to address class imbalance by re-weighting loss contributions based on effective sample counts. For each class $y$ with $n_y$ training examples, the effective number accounts for information overlap:
\begin{equation}
\text{EN}_y = \frac{1-\beta^{n_y}}{1-\beta}, \quad \text{where } \beta=0.999
\end{equation}
Raw class weights are computed as $w_y^{\text{raw}} = 1/\text{EN}_y$. These are normalized to unit mean:
\begin{equation}
w_y^{\text{norm}} = w_y^{\text{raw}} \cdot \frac{C}{\sum_{c=1}^C w_c^{\text{raw}}}
\end{equation}
To prevent destabilization from extreme weight ratios, we apply clamping:
\begin{equation}
w_y = \text{clip}(w_y^{\text{norm}}, w_{\min}, w_{\max})
\end{equation}
with $w_{\min}=0.25$ and $w_{\max}=4.0$, followed by re-normalization to unit mean. Hence the loss function is formulated as follows:
\begin{equation}
\mathcal{L}_{\text{CB-CE}}(x,y) = -w_y \sum_{c=1}^C \tilde{y}_c \log \text{softmax}(\mathbf{z}(x))_c
\end{equation}
\paragraph{Temperature Scaling}
After training, we calibrate model confidence using temperature scaling to improve probability estimates. We fit a single temperature parameter $T$ by minimizing negative log-likelihood on the development set:
\begin{equation}
\begin{split}
T^\star &= \arg\min_T \sum_{(x,y)\in\mathcal{D}_{\mathrm{dev}}}
\Bigg[-\frac{z_y(x)}{T} \\
&\qquad\qquad\quad + \log\!\sum_{c=1}^C
\exp\!\left(\frac{z_c(x)}{T}\right)\Bigg]
\end{split}
\end{equation}

Calibrated probabilities are then computed as:
\begin{equation}
p_\theta^T(y \mid x) = \text{softmax}(\mathbf{z}(x)/T^\star)_y
\end{equation}
Temperature scaling is particularly important for binary classification as it corrects systematic over-confidence or under-confidence in model predictions without requiring retraining.

\paragraph{Decision Threshold Tuning}
Threshold $t$ is tuned on calibrated probabilities to maximize development set macro-F1:
\begin{align}
t^\star = \operatorname*{arg\,max}_{t\in[0,1]}\;
\mathrm{F1}_{\mathrm{macro}}\big(\hat{\mathbf{y}}_t\big),\;
\hat{\mathbf{y}}_t \!=\! \big(\hat{y}_t(x)\big)_{x\in\mathcal{D}_{\mathrm{dev}}}
\end{align}

This approach balances precision and recall across both classes, particularly important when minority class performance is critical. In practice, $t^\star$ typically falls in the range $[0.3, 0.6]$ depending on class imbalance severity.

\subsubsection{Multi-Class Classification Optimizations}\label{multi}

For multi-class settings with long-tailed distributions (Tasks B and C), we adopt Class-Balanced Focal Loss, which combines effective-number weighting with focal loss modulation. The CB-Focal loss for a sample $(x,y)$ is:
\begin{equation}
\mathcal{L}_{\text{CB-Focal}}(x,y) = - w_y (1-p_y)^\gamma \log(p_y)
\end{equation}

where:
\begin{itemize}
    \item $w_y$ is the CB weight for class $y$ (computed as in the previous section)
    \item $p_y$ is the predicted probability for the true class
    \item $\gamma$ is the focusing parameter (typically $\gamma \in [1,3]$, we use $\gamma=2$)
\end{itemize}

The $(1-p_y)^\gamma$ term down-weights easy examples (high $p_y$) while amplifying hard or misclassified ones (low $p_y$). This dual mechanism:
\begin{itemize}
    \item Re-weights by class frequency (via $w_y$)
    \item Re-weights by example difficulty (via $(1-p_y)^\gamma$)
\end{itemize}

This loss is particularly effective for Tasks B and C, where rare subcategories (e.g., \textit{backhanded compliments}, \textit{condescending explanations}) are overwhelmed by frequent classes (\textit{derogation}). By combining class re-weighting and hard-example mining, CB-Focal improves macro-F1 by 3--5 points, ensuring minority classes are not neglected while reducing over-confidence on majority classes.

\paragraph{Class-Aware Batching.} Standard uniform sampling produces batches dominated by majority classes (e.g., 12--14 majority, 2--4 minority instances), leading to inefficient learning on minority classes. Given per-device batch size $B$ and $C$ classes, the quota per class is:
\begin{equation}
k = \lfloor B/C \rfloor
\end{equation}

For Task B with $C=4$ classes and $B=16$, we sample $k=4$ instances per class per batch. For each training batch:
\begin{enumerate}
    \item Partition training data into class-specific subsets $\mathcal{D}_1, \ldots, \mathcal{D}_C$
    \item For each class $c \in \{1,\ldots,C\}$, randomly sample $k$ instances from $\mathcal{D}_c$ with replacement
    \item Concatenate samples to form batch $\mathcal{B}$ of size $k \cdot C \leq B$
    \item Shuffle combined batch to avoid positional biases
\end{enumerate}
This ensures minority classes contribute equally despite smaller pool sizes. Without replacement would exhaust small classes quickly, leading to epoch-boundary effects.

\subsubsection{Training Hyperparameters}\label{hyper}

Table~\ref{tab:hyperparameters} summarizes the hyperparameters selected across tasks.

\begin{table}[h]
\centering
\setlength{\tabcolsep}{0.5pt}
\small
\begin{tabular}{@{}lccc@{}}
\toprule
\textbf{Hyperparameter} & \textbf{Task 1.1/Task A} & \textbf{Task B} & \textbf{Task C}  \\
\midrule
Learning rate & 2$\times10^{-4}$ & 6$\times10^{-5}$ & 2$\times10^{-5}$  \\
Batch size & 16 & 16 & 16  \\
Gradient accumulation & 2 & 2 & 2  \\
Training epochs & 5 & 8 & 12  \\
Warmup ratio & 0.1 & 0.1 & 0.1  \\
Weight decay & 0.01 & 0.01 & 0.01  \\
Max sequence length & 512 & 512 & 512  \\
\midrule
Label smoothing ($\epsilon$) & 0.05 & 0.05 & 0.05 \\
CB-CE $\beta$ & 0.999 & 0.999 & 0.999  \\
CB-CE $w_{\min}$ & 0.25 & 0.25 & 0.25  \\
CB-CE $w_{\max}$ & 4.0 & 4.0 & 4.0  \\
Focal loss $\gamma$ & --- & 2.0 & 2.0 \\
\midrule
LoRA rank ($r$) & 32 & 96 & 96  \\
LoRA alpha ($\alpha$) & 64 & 192 & 192  \\
LoRA dropout & 0.1 & 0.2 & 0.2  \\
\bottomrule
\end{tabular}
\caption{Hyperparameter configuration across tasks.}
\label{tab:hyperparameters}
\end{table}

\subsection{Zero-shot Baselines} \label{zeroshot}
For the zero-shot baselines, LLMs including \textsc{Qwen2.5-72B-Instruct}, \textsc{LLaMA-3.3-70B-Instruct}, and \textsc{Cogito-70B} were evaluated using task-specific simple prompts. Each model was queried with a zero-shot classification instruction tailored to the corresponding EDOS task as follows:

\begin{itemize}
  \item \textbf{EXIST Task 1.1/ EDOS Task A}: \textit{``Classify the following text as sexist or not sexist.''}
 \item \textbf{Task B}: \textit{``Classify the following text into \textbf{one} of the sexism categories:
  1) Threats, plans to harm and incitement;
  2) Derogation;
  3) Animosity;
  4) Prejudiced discussions.''}
  \item \textbf{Task C}: 
  \textit{``Classify the following text into \textbf{one} of the sexism subcategories:
  1.1) Threats of harm;
  1.2) Incitement and encouragement of harm;
  2.1) Descriptive attacks;
  2.2) Aggressive and emotive attacks;
  2.3) Dehumanising attacks \& overt sexual objectification;
  3.1) Casual use of gendered slurs, profanities, and insults;
  3.2) Immutable gender differences and gender stereotypes;
  3.3) Backhanded gendered compliments;
  3.4) Condescending explanations or unwelcome advice;
  4.1) Supporting mistreatment of individual women;
  4.2) Supporting systemic discrimination against women as a group.''}
\end{itemize}

\subsection{Personas Progressive Prompt Design}\label{personas}
Our progressive prompt construction method incrementally integrates role-conditioning, reasoning mechanisms, expert knowledge, and guideline refinements into our prompt design approach. Each iteration aims at improving performance, interpretability, and alignment with human annotator behavior as listed below:

\begin{itemize}
    \item \(\mathcal{P}_1\):
    A basic prompt with few-shot examples and role identity. This sets the task framing and label expectations, testing the model's generalization capabilities.

    \item\(\mathcal{P}_2\):
    Enhances reasoning by expanding the role identity and background into a specialized expert persona. This stage encourages more structured reasoning by clearly separating the task into sub-steps,
    \item\(\mathcal{P}_3\):
    Integrates formal sexism definition aligned with annotation guidelines.

    \item\(\mathcal{P}_4\):
    Adds nuanced and multilingual examples to strengthen generalization.

    \item\(\mathcal{P}_5\):
       Based on manual error analysis, the prompt incorporates refined guidelines, including considerations of the tweet author's intent and target audience, as well as the inclusion of edge cases (e.g., profanity and slang) to guide borderline decisions (examples provided in Table \ref{tab:prompt_design}).
\end{itemize}

\begin{table}[h]
\setlength{\tabcolsep}{1mm}
\centering
\begin{tabular}{cp{7cm}}
\hline
 \# & Example \\
\hline
\(\mathcal{P}_1\) &  You are a psychologist. Classify the tweets as sexist or not sexist. Here are some examples... \\

\(\mathcal{P}_2\) & + (45 y/o female, Argentina, Ph.D. Psychology) Focus on language, cognitive biases, and the psychological effects of sexism. (1) Analyze the text carefully; (2) Think before responding; (3) Classify the tweet as sexist (1) or not sexist (0). \\

\(\mathcal{P}_3\) & + Sexism is gender-based prejudice, stereotyping, or discrimination, typically against women. Label as sexist (1) if the tweet: (a) is sexist itself, (b) describes a sexist situation, or (c) criticizes sexist behavior. \\

\(\mathcal{P}_4\) & + "She got promoted because they needed 'more women in leadership.'" → Sexist (1). \newline "No tengo nada contra las mujeres, pero en cargos altos siempre rinden menos." → Sexist (1). \\

\(\mathcal{P}_5\) & + (1) Consider the author's intent (insult, joke, venting, shaming). (2) Profanity alone does not indicate sexism — consider the context carefully. \\
\hline
\end{tabular}
\caption{Examples illustrating the structure and progression of each prompt stage.}
\label{tab:prompt_design}
\end{table}

The selected personas are prompted with unique descriptions as shown in Table~\ref{tab:unified_personas}. The six expert personas used in our classification framework are described, each representing a distinct interpretive lens on sexism. These roles were carefully constructed to reflect diverse perspectives, including legal, psychological, linguistic, and lived experience. To increase realism, demographic and educational attributes were heuristically aligned with annotator metadata from the EXIST 2025 dataset, mimicking the diversity and subjectivity found in real-world human annotation.

\begin{table*}[t]
\centering
\begin{adjustbox}{max width=0.98\textwidth}
\setlength{\tabcolsep}{2mm}
\begin{tabular}{@{}p{3cm}p{3.5cm}p{10cm}@{}}
\toprule
\textbf{Persona} & \textbf{Perspective} & \textbf{Role Description (Prompt)} \\
\midrule

Layperson 
& General public perspective 
& An average person (18–22 y/o male, Portugal, B.A.) with no specialized expertise in law, psychology, or linguistics. \\

Linguist 
& Language patterns and bias 
& A linguist (23–45 y/o male, Poland, B.A. Linguistics) specializing in semantics, pragmatics, and discourse analysis, with a focus on gendered language. \\

Psychologist 
& Emotional and cognitive effects 
& A psychologist specializing in language, cognitive biases, and the psychological effects of sexism. \\

Legal Expert 
& Legal view on gender-based discrimination 
& A legal expert (46+ y/o male, Portugal, M.A. Law) specializing in anti-discrimination laws, workplace regulations, and gender equality. \\

Gender Expert 
& Structural and ideological analysis 
& A gender studies expert (46+ y/o female, UK, B.A. Gender Studies) with deep knowledge of gender theories, power dynamics, and social structures. \\

Sexism Victim 
& Personal impact of experiencing sexism 
& A person (18–22 y/o female, South Africa, H.S. diploma) who has personally experienced sexism and understands its emotional and social impact. \\

\bottomrule
\end{tabular}
\end{adjustbox}
\caption{Persona perspectives and detailed role descriptions.}
\label{tab:unified_personas}
\end{table*}

\subsubsection{Preliminary Investigation}

Table~\ref{tab:appendix_persona_prelim} reports a preliminary investigation conducted on relatively smaller-scale LLMs (7--8B parameters), namely \textsc{Mistral-7B}, \textsc{Dolphin3-8B}, and\textsc{Cogito-7B}. The table compares the baseline prompt (``Classify the given text as sexist or not sexist'') against our first enhanced persona-based prompt, denoted as $\mathcal{P}_1$, evaluated on the EXIST 2025 development set. In contrast to the baseline, $\mathcal{P}_1$ incorporates few-shot examples and an explicit role identity (e.g., psychologist, linguist, normal person) to guide model reasoning.  

Overall, the results confirm that even modest prompt enhancements yield consistent improvements across metrics. For \textsc{Mistral-7B} and\textsc{Cogito-7B}, $\mathcal{P}_1$ produced reliable gains in ICM, normalized ICM, and F1, while \textsc{Dolphin3-8B}, despite its weak baseline performance, exhibited the largest relative improvements. Importantly, the absolute best scores are consistently achieved by\textsc{Cogito-7B}, suggesting that stronger base models are less sensitive to prompt design but still benefit incrementally. In contrast, weaker models (e.g., \textsc{Dolphin3-8B}) depend more heavily on the additional structure introduced by persona framing.  

Finally, performance varied across personas: roles such as Normal Person and Legal Studies Expert saw more pronounced gains, while Sexism Victim and Linguist produced only modest improvements. These exploratory findings were not used as final results but rather as guidance for subsequent large-scale experiments, helping us refine prompt design choices and identify promising personas for sexism detection.

\begin{table*}[h]
\small
\setlength{\tabcolsep}{1mm}
\centering
\begin{tabular}{@{}llccccccccc@{}}
\toprule
\textbf{Persona} & \textbf{Model} 
& \textbf{ICM$_{1}$} & \textbf{ICM$_{2}$} & \textbf{ICM\%↑} 
& \textbf{Norm$_{1}$} & \textbf{Norm$_{2}$} & \textbf{Norm\%↑} 
& \textbf{F1$_{1}$} & \textbf{F1$_{2}$} & \textbf{F1\%↑} \\
\midrule
\multirow{3}{*}{Psychologist} 
  & \textsc{Mistral-7B}   & 0.2388 & 0.4059 & +69.96\% & 0.6195 & 0.7031 & +13.49\% & 0.7433 & 0.8019 & +7.89\% \\
  & \textsc{Dolphin3-8B}  & -0.2299 & -0.2299 & 0.00\% & 0.3850 & 0.3850 & 0.00\% & 0.5245 & 0.5245 & 0.00\% \\
  &\textsc{Cogito-7B}    & 0.4230 & 0.4703 & +11.16\% & 0.7116 & 0.7353 & +3.33\% & 0.8052 & 0.8233 & +2.25\% \\
\midrule

\multirow{3}{*}{Sexism Victim}
  & \textsc{Mistral-7B}   & -0.0258 & 0.3559 & +1478.68\% & 0.4871 & 0.6781 & +39.16\% & 0.6672 & 0.7882 & +18.13\% \\
  & \textsc{Dolphin3-8B}  & -0.2265 & -0.1298 & +42.70\% & 0.3867 & 0.4351 & +12.51\% & 0.5244 & 0.5866 & +11.86\% \\
  &\textsc{Cogito-7B}    & 0.4089 & 0.4505 & +10.16\% & 0.7046 & 0.7254 & +2.95\% & 0.8024 & 0.8148 & +1.54\% \\
\midrule

\multirow{3}{*}{Linguist}
  & \textsc{Mistral-7B}   & 0.1882 & 0.3435 & +82.56\% & 0.5941 & 0.6718 & +13.11\% & 0.7327 & 0.7800 & +6.44\% \\
  & \textsc{Dolphin3-8B}  & -0.3184 & -0.3184 & 0.00\% & 0.3407 & 0.3407 & 0.00\% & 0.4724 & 0.4724 & 0.00\% \\
  &\textsc{Cogito-7B}    & 0.3967 & 0.4055 & +2.22\% & 0.6985 & 0.7029 & +0.63\% & 0.7957 & 0.8011 & +0.68\% \\
\midrule

\multirow{3}{*}{Legal Studies Expert}
  & \textsc{Mistral-7B}   & 0.2317 & 0.3586 & +54.75\% & 0.6159 & 0.6794 & +10.30\% & 0.7494 & 0.7858 & +4.86\% \\
  & \textsc{Dolphin3-8B}  & -0.2300 & 0.0427 & +118.57\% & 0.3849 & 0.5214 & +35.45\% & 0.5261 & 0.6693 & +27.26\% \\
  &\textsc{Cogito-7B}    & 0.4375 & 0.4772 & +9.08\% & 0.7189 & 0.7387 & +2.75\% & 0.8117 & 0.8254 & +1.69\% \\
\midrule

\multirow{3}{*}{Normal Person}
  & \textsc{Mistral-7B}   & 0.3445 & 0.3952 & +14.72\% & 0.6723 & 0.6977 & +3.78\% & 0.7816 & 0.7985 & +2.16\% \\
  & \textsc{Dolphin3-8B}  & -0.2395 & -0.0675 & +71.82\% & 0.3802 & 0.4662 & +22.62\% & 0.5212 & 0.6213 & +19.20\% \\
  &\textsc{Cogito-7B}    & 0.3982 & 0.4863 & +22.11\% & 0.6992 & 0.7433 & +6.31\% & 0.7978 & 0.8287 & +3.87\% \\
\midrule

\multirow{3}{*}{Gender Studies Expert}
  & \textsc{Mistral-7B}   & 0.2875 & 0.3956 & +37.57\% & 0.6438 & 0.6979 & +8.41\% & 0.7590 & 0.7986 & +5.22\% \\
  & \textsc{Dolphin3-8B}  & -0.2493 & -0.0769 & +69.15\% & 0.3753 & 0.4615 & +22.99\% & 0.5188 & 0.6164 & +18.78\% \\
  &\textsc{Cogito-7B}    & 0.3826 & 0.4339 & +13.41\% & 0.6914 & 0.7171 & +3.72\% & 0.7930 & 0.8098 & +2.12\% \\
\bottomrule
\end{tabular}
\caption{Preliminary results of persona-based prompts, comparing baseline prompt with \(\mathcal{P}_1\). Metrics include ICM, normalized ICM, and F1, with relative improvements (\%↑).}
\label{tab:appendix_persona_prelim}
\end{table*}

\subsection{Collaborative Expert Judgment (CEJ) Algorithm and Complexity}
Algorithm~\ref{alg:persona-debate} summarizes the CEJ pipeline for a dataset of $N$ instances and a persona set of size $K$. CEJ requires $K$ persona inferences, one discussion, one summarization, and one judgment per instance, yielding $N \times (K+3)$ total LLM calls with complexity $O(NK)$. The cost therefore scales linearly with both the dataset size and the number of personas. \footnote{The analysis below concerns the interaction-level complexity of the CEJ procedure rather than low-level runtime or token usage.}
\begin{algorithm}[h]
\caption{Collaborative Expert Judgment (CEJ)}
\label{alg:persona-debate}
\textbf{Input}: Dataset $\mathcal{D} = \{(x_i, \text{id}_i)\}_{i=1}^{N}$, persona set $\mathcal{Y} = \{y_1, \dots, y_K\}$, Large Language Model $\mathcal{M}$ \\
\textbf{Output}: Opinion matrix $\mathcal{O} = \{O_{i,k}\}$, summary $S_i$, and final judgment $J_i$ for each $x_i$

\begin{algorithmic}[1]
\FOR{each instance $(x_i, \text{id}_i) \in \mathcal{D}$}
    \STATE \textbf{Initial Persona Classification}
    \FOR{each persona $y_k \in \mathcal{Y}$}
        \STATE Construct persona-specific classification prompt $\mathcal{P}_{\text{cls}}(x_i, y_k)$
        \STATE $O_{i,k} \gets \mathcal{M}.\texttt{invoke}(\mathcal{P}_{\text{cls}})$
    \ENDFOR
    
    \STATE \textbf{Simulated Panel Discussion}
    \STATE Construct discussion prompt $\mathcal{P}_{\text{disc}}(x_i, \{O_{i,k}\}_{k=1}^K, \mathcal{Y})$
    \STATE $\mathcal{D}_i \gets \mathcal{M}.\texttt{invoke}(\mathcal{P}_{\text{disc}})$
    
    \STATE \textbf{Summary Generation}
    \STATE Construct summarization prompt $\mathcal{P}_{\text{sum}}(\mathcal{D}_i, \mathcal{Y})$
    \STATE $S_i \gets \mathcal{M}.\texttt{invoke}(\mathcal{P}_{\text{sum}})$
    
 \STATE \textbf{Final Judgment with Justification}
\STATE Construct judgment prompt $\mathcal{P}_{\text{judge}}(x_i, \{O_{i,k}\}_{k=1}^K, S_i)$
\STATE $J_i \gets \mathcal{M}.\texttt{invoke}(\mathcal{P}_{\text{judge}})$

\ENDFOR
\end{algorithmic}
\end{algorithm}

\subsection{Qualitative Analysis}

\begin{figure*}[h]
  \centering
    \includegraphics[width=0.9\linewidth]{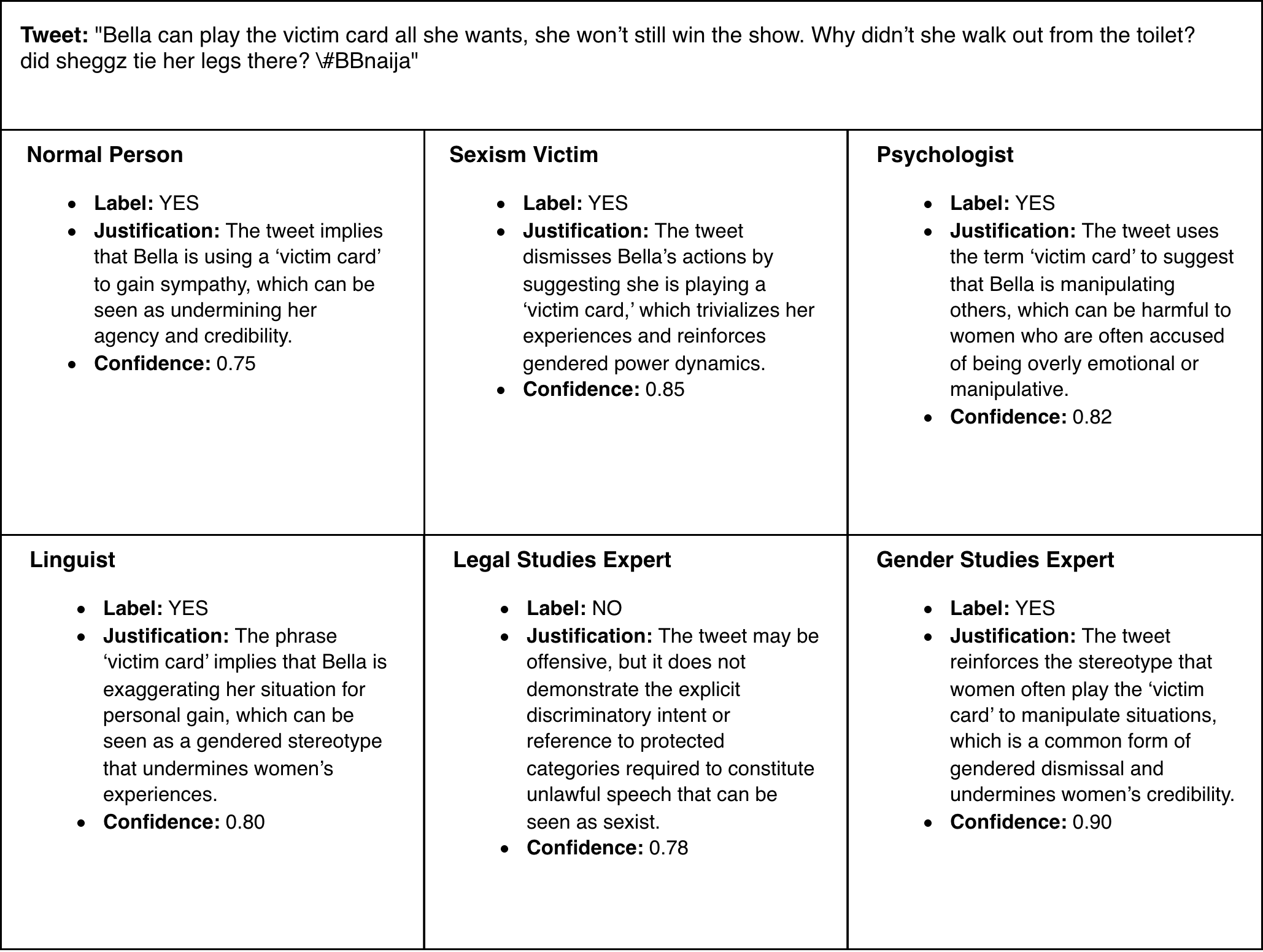}
  \caption{An example output from the initial classification stage of the CEJ framework.}
  \label{fig:initial_example}
\end{figure*}

\begin{figure*}[h]
  \centering
    \includegraphics[width=0.9\linewidth]{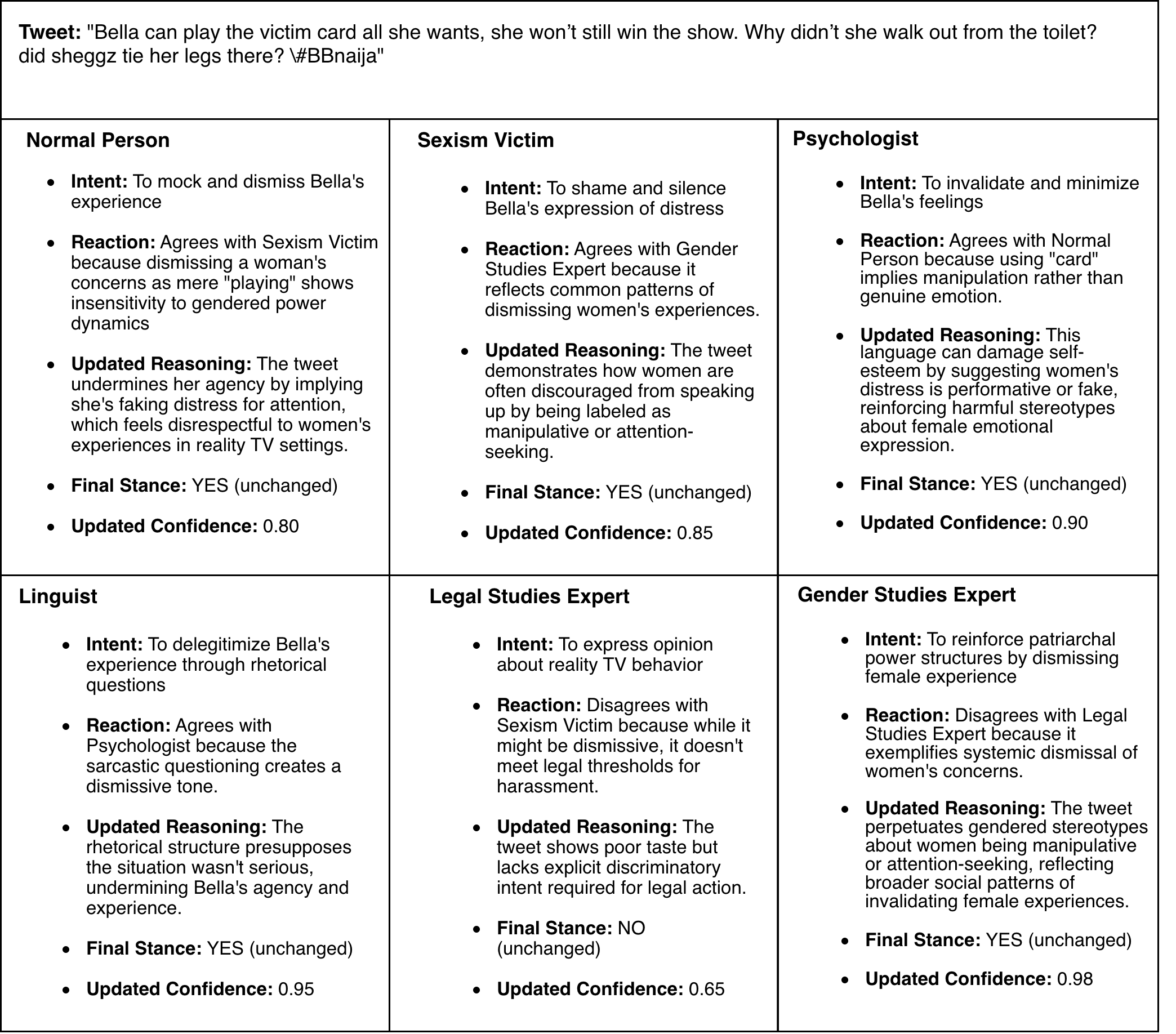}
  \caption{An example output from the structured debate stage of the CEJ framework.}
  \label{fig:cej_example}
\end{figure*}

Figures~\ref{fig:initial_example} and~\ref{fig:cej_example} illustrate the 
Collaborative Expert Judgment (CEJ) framework using a representative tweet 
from EXIST 2025. Figure~\ref{fig:initial_example} shows the initial 
classification stage, where six expert personas independently assess the 
input. Each provides a binary label (\texttt{YES}/\texttt{NO}), a 
domain-grounded justification, and a confidence score. Most personas classify 
the tweet as sexist, citing trivialization of female emotional expression or 
reinforcement of gendered stereotypes; only the \textit{Legal Expert} 
disagrees, arguing the statement lacks discriminatory intent under legal 
criteria.

Figure~\ref{fig:cej_example} presents the structured debate stage, where 
personas reflect on others' reasoning and may revise their stance and 
confidence. This deliberative process demonstrates how LLM-based agents 
engage in structured argumentation—aligning with or challenging one another 
based on shared evidence and domain-specific reasoning. While most personas 
maintain their original \texttt{YES} classification, their justifications 
become more nuanced and confidence scores are recalibrated. The 
\textit{Legal Expert} again maintains \texttt{NO}, citing insufficient legal 
grounds.

\begin{figure}[tb]
  \centering
  \includegraphics[width=\linewidth]{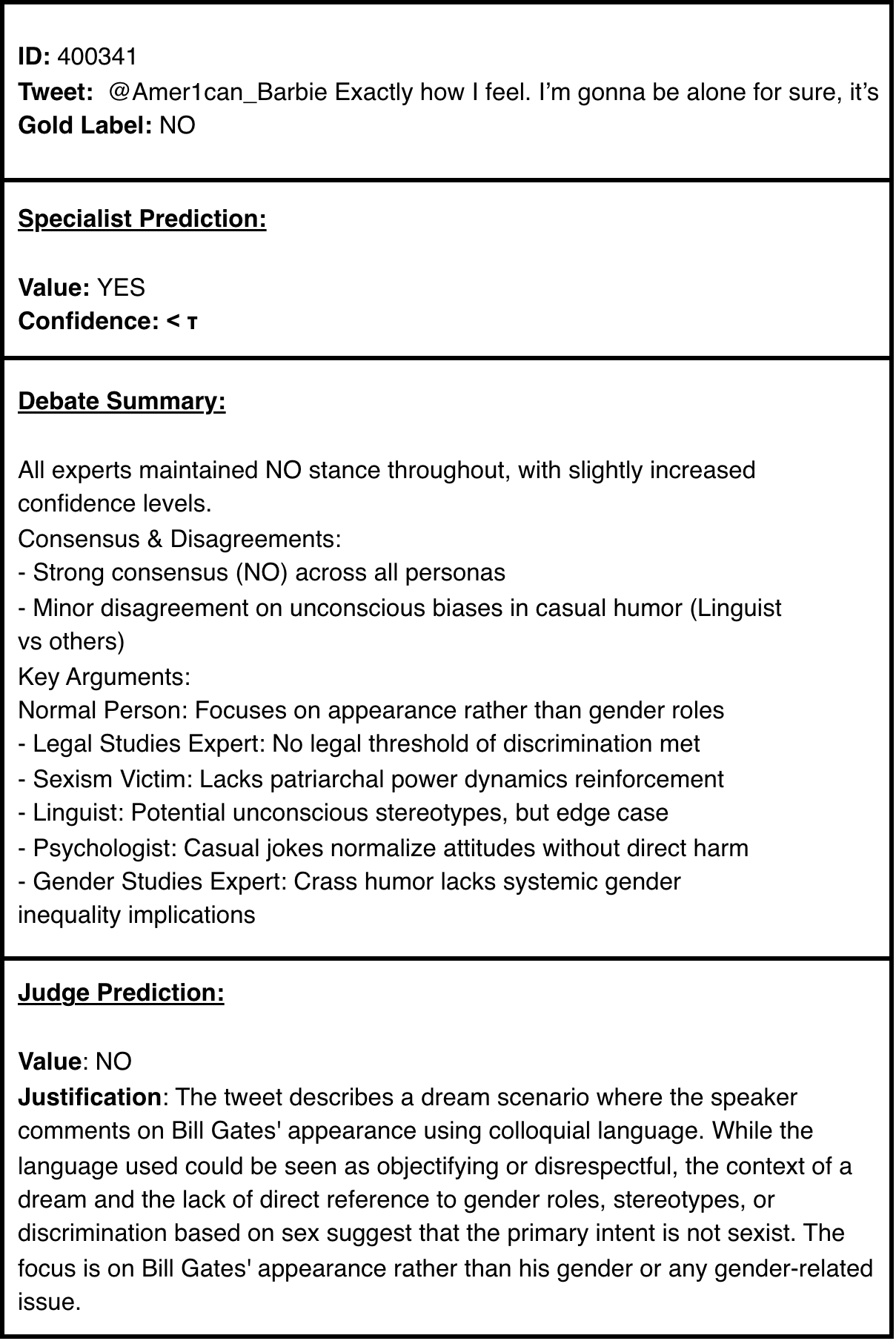}
  \caption{Error correction via confidence-based routing. The specialist's 
  false positive (confidence $< \tau$) is overridden by CEJ deliberation.}
  \label{fig:routing_correction}
\end{figure}

Figure~\ref{fig:routing_correction} illustrates successful error correction 
via confidence-based routing. Here, the specialist misclassifies the input as 
\texttt{YES} (sexist) with confidence below threshold $\tau$, triggering CEJ 
escalation. The expert panel classifies it as \texttt{NO}, reasoning that the 
text targets physical appearance rather than gender, lacks patriarchal power 
dynamics, and occurs in a casual context. The judge concurs, overriding the 
false positive. This case demonstrates how routing captures borderline errors: 
low-confidence predictions are delegated to multi-perspective deliberation, 
where richer contextual and intent analysis corrects specialist mistakes.


\section{Prompts}
This section presents the prompt templates used throughout the CEJ framework. Initial persona judgments are collected using the prompt shown in Figure~\ref{Persona-CEJ}. This is followed by the structured debate prompt (Figure~\ref{Persona-Discussion-CEJ}), which enables each persona to reflect on peer opinions, revise its stance if persuaded, and recalibrate its confidence. The discussion is then summarized and passed to the judge prompt (Figure~\ref{judge_cej}), which concludes the process by generating a final label along with a justification and calibrated confidence score. All prompts rely on consistent definitions and structured output formats (e.g., JSON objects with fields such as \texttt{label}, \texttt{justification}, and \texttt{confidence}) to support interpretability and downstream evaluation.

\begin{figure*}
    
\begin{tcolorbox}[colback=lightgray!5, colframe=black!80, sharp corners=south,  boxrule=0.4pt]
You are \texttt{\{personas\_description\}} tasked with classifying the following tweet for sexism. \\
\textbf{Tweet:} \texttt{{\{tweet\_text\}}} \\

\textbf{Your task:}
\begin{enumerate}
  \item Read the guidelines below carefully
  \item Analyze the tweet carefully for sexism.
  \item Think before responding.
  \item Decide the final label: 1 (sexist) or 0 (not sexist).
  \item Provide a short justification for their label based on their role.
  \item Output a confidence score between 0.0 and 1.0 reflecting your certainty.
\end{enumerate}

\textbf{Sexism Definition} \\
\texttt{\{definition\}}\\
\textbf{Objective:} \\
\texttt{\{objective\}}\\
\textbf{Here are some examples:} \\
\texttt{\{examples\}}\\
\textbf{Output Example:}  Provide only a valid JSON object like the following example:
\begin{lstlisting}[style=jsonstyle]
{
"persona": "Normal Person",
"label": "1",
"justification": "The tweet stereotypes women's intelligence.",
"confidence": "0.87"
}
\end{lstlisting}

\end{tcolorbox}
\caption{Template for initial persona classification in the CEJ framework.}
\label{Persona-CEJ}
\end{figure*}

\begin{figure*}
    
\begin{tcolorbox}[colback=lightgray!5, colframe=black!80, sharp corners=south,  boxrule=0.4pt]
You are continuing the expert panel discussion on the following tweet:\\
\textbf{Tweet:} \texttt{{\{tweet\_text\}}} \\
\textbf{Initial Opinions:} \texttt{{\{persona\_opinions\}}}
\\
Now, each persona must:
\begin{enumerate}
  \item Read all other personas' initial opinions.
  \item Reflect on whether their own reasoning is still the strongest.
  \item Engage with at least one other persona by agreeing or disagreeing with their argument.
  \item Update their stance if persuaded, or affirm their original decision.
  \item Reassess and adjust their confidence accordingly.
\end{enumerate}
\textbf{Sexism Definition} \\
\texttt{\{definition\}}\\
\textbf{Objective:} \\
\texttt{\{objective\}}\\
\textbf{Here are some examples:} \\
\texttt{\{examples\}}\\
\textbf{Important Notes:}
\begin{itemize}
  \item Confidence can be increased if supported by solid reasoning, or reduced if uncertainty arises.
  \item Final answers must state if the stance is changed or unchanged.
\end{itemize}

\textbf{Output Example} (per persona):
\begin{lstlisting}[style=jsonstyle]
{
  "persona": "Sexism Victim",
  "intent": "The author's intent is to shame the woman by dismissing her distress.",
  "reaction": "Agree with Linguist because their interpretation highlights the use of gendered stereotypes.",
  "updated_reasoning": "While my initial view focused on tone, I now realize the tweet uses the 'victim card' trope to discredit women's emotional responses.",
  "final_stance": 1 (changed from 0),
  "updated_confidence": 0.72
}
\end{lstlisting}
\end{tcolorbox}
\caption{Structured discussion prompt for reasoning and stance revision in the CEJ framework.}
\label{Persona-Discussion-CEJ}
\end{figure*}

\begin{figure*} 
\begin{tcolorbox}[colback=lightgray!5,  colframe=black!80, sharp corners=south,  boxrule=0.4pt]
You are an impartial judge of the detection of sexist language. Review the following expert panel summary and the original tweet being discussed then classify it for sexism. 

You have the following information:\\
\textbf{Tweet:} \texttt{{\{tweet\_text\}}} \\
\textbf{Discussion summary:}\\ \texttt{{\{summary\}}}

\textbf{Your task:}
\begin{enumerate}
 \item Read the guidelines below carefully
  \item Review  expert panel summary.
  \item Analyze the tweet for sexism.
  \item Think before responding.
  \item Decide the final label: 1 (sexist) or 0 (not sexist).
  \item  Provide a brief justification for your decision.
\item State your confidence score (0.0–1.0).

\end{enumerate}

\textbf{Sexism Definition:} \\
\texttt{\{definition\}}\\
\textbf{Objective:} \\
\texttt{\{objective\}}\\
\textbf{Here are some examples:} \\
\texttt{\{examples\}}\\
\textbf{Output Example:} Provide only a valid JSON object like the following:

\begin{lstlisting}[style=jsonstyle]
{
  "label": 0,
  "justification": "Although the tweet contains profanity and strong language, it is not clearly directed at women. Based on the personas' disagreement and the classification guidelines, the tweet is offensive but not sexist.",
  "confidence": 0.79
}
\end{lstlisting}
\end{tcolorbox}
\caption{Prompt for the judge revision following persona discussion in the CEJ method.}
\label{judge_cej}
\end{figure*}

\end{document}